\documentclass[10pt,twocolumn,letterpaper]{article}

\usepackage{cvpr}
\usepackage{times}
\usepackage{epsfig}
\usepackage{graphicx}
\usepackage{amsmath}
\usepackage{amssymb}
\usepackage{bm}
\usepackage{algorithm}
\usepackage{algorithmic}
\usepackage{wrapfig}
\usepackage[breaklinks=true,bookmarks=false]{hyperref}

\cvprfinalcopy % *** Uncomment this line for the final submission

\def\cvprPaperID{****} % *** Enter the CVPR Paper ID here

\ifcvprfinal\fi
\begin{document}

%%%%%%%%% TITLE
\title{Deep Regression Forests for Age Estimation}

\author{Wei Shen$^{1,2}$, Yilu Guo$^1$, Yan Wang$^2$, Kai Zhao$^3$, Bo Wang$^4$, Alan Yuille$^2$\\
$^1$ Key Laboratory of Specialty Fiber Optics and Optical Access Networks, \\Shanghai Institute for Advanced Communication and Data Science, \\School of Communication and Information Engineering, Shanghai University\\
$^2$ Department of Computer Science, Johns Hopkins University\\ 
$^3$ College of Computer and Control Engineering, Nankai University  $^4$ Hikvision Research\\
{\tt\footnotesize
\{shenwei1231,gyl.luan0,wyanny.9,zhaok1206,wangbo.yunze,alan.l.yuille\}@gmail.com}}

\maketitle
\thispagestyle{empty}

%%%%%%%%% ABSTRACT
\begin{abstract}
Age estimation from facial images is typically cast as a nonlinear regression problem. The main challenge of this problem is the facial feature space w.r.t. ages is heterogeneous, due to the large variation in facial appearance across different persons of the same age and the non-stationary property of aging patterns. In this paper, we propose Deep Regression Forests (DRFs), an end-to-end model, for age estimation. DRFs connect the split nodes to a fully connected layer of a convolutional neural network (CNN) and deal with heterogeneous data by jointly learning input-dependant data partitions at the split nodes and data abstractions at the leaf nodes. This joint learning follows an alternating strategy: First, by fixing the leaf nodes, the split nodes as well as the CNN parameters are optimized by Back-propagation; Then, by fixing the split nodes, the leaf nodes are optimized by iterating a step-size free and fast-converging update rule derived from Variational Bounding. We verify the proposed DRFs on three standard age estimation benchmarks and achieve state-of-the-art results on all of them.
\end{abstract}

%%%%%%%%% BODY TEXT
\section{Introduction}\label{sec:intro}

There has been a growing interest in age estimation from facial images, driven by the increasing demands for a variety of potential applications in forensic research~\cite{Ref:Alkass10}, security control~\cite{Ref:HanOJ13}, human-computer interaction (HCI)~\cite{Ref:HanOJ13} and social media~\cite{Ref:Rothe16}. Although this problem has been extensively studied, the ability to automatically estimate ages accurately and reliably from facial images is still far from meeting human performance.

\begin{figure}[!t]
\centering
\includegraphics[trim=0cm 6cm 0cm 7cm, clip=true, width=1.0\linewidth]{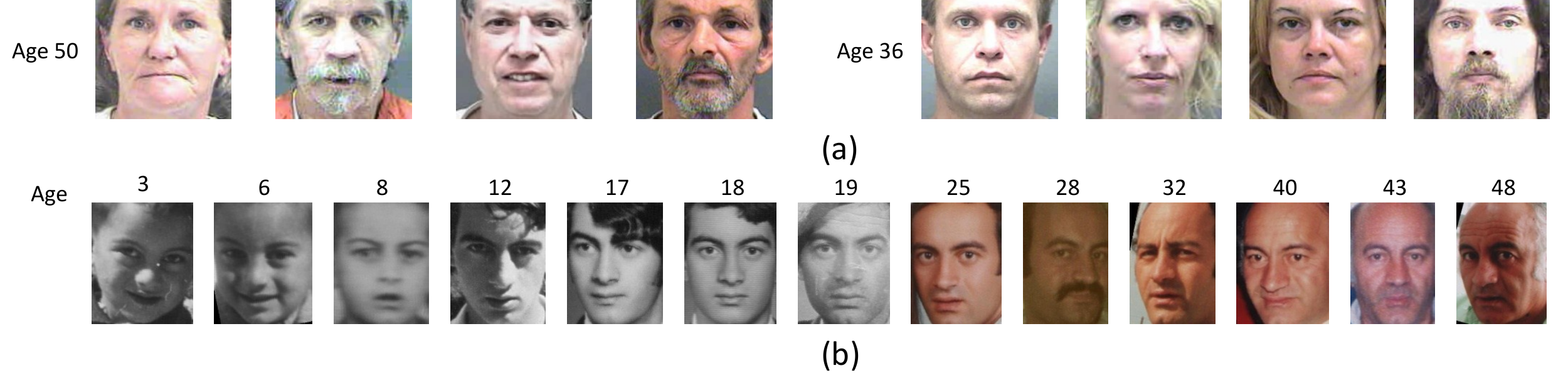}
\caption{(a) The large variation in facial appearance across different persons of the same age. (b) Facial images of a person from childhood to adulthood. Note that, Facial aging effects appear as changes in the shape of the face during childhood and changes in skin texture during adulthood, respectively.}\label{fig:Example}
\end{figure}

There are two kinds of age estimation tasks. One is real age estimation, which is to estimate the precise biological (chronological) age of a person from his or her facial image; the other is age group estimation~\cite{Ref:LeviH15}, which is to predict whether a person's age falls within some range rather than predicting the real chronological age. In this paper, we focus on the first task, i.e., precise age regression. To address this problem, the key is to learn a nonlinear mapping function between facial image features and the real chronological age. However, to learn such a mapping is challenging. The main difficulty is the facial feature space w.r.t ages is heterogeneous, due to two facts: 1) there is a large variation in facial appearance across different persons of the same age (Fig.~\ref{fig:Example}(a)); 2) the human face matures in different ways at different ages, e.g., bone growth in childhood and skin wrinkles in adulthood~\cite{Ref:RamanathanCB09} (Fig.~\ref{fig:Example}(b)).

To model such heterogeneous data, existing age estimation methods either find a kernel-based global non-linear mapping~\cite{Ref:GuoMFH09,Ref:GuoM11}, or apply divide-and-conquer strategies to partition the data space and learn multiple local regressors~\cite{Ref:HanOLJ15}. However, each of them has drawbacks: Learning non-stationary kernels is inevitably biased by the heterogeneous data distribution and thus easily causes over-fitting~\cite{Ref:ChangCH11}; Divide-and-conquer is a good strategy to learn the non-stationary age changes in human faces, but the existing methods make hard partitions according to ages~\cite{Ref:HaraC14,Ref:HanOLJ15}. Consequently, they may not find homogeneous subsets for learning local regressors~\cite{Ref:Huang17}.

To address the above-mentioned challenges, we propose differentiable regression forests for age estimation. Random forests or randomized decision trees~\cite{Ref:Amit97,Ref:Breiman01,Ref:Shotton13}, are a popular ensemble predictive model, in which each tree structure naturally performs data partition at split nodes and data abstraction at leaf nodes~\cite{Ref:Shen12}. Traditional regression forests make hard data partitions, based on heuristics such as using a greedy algorithm where locally-optimal hard decisions are made at each split node~\cite{Ref:Amit97}. Unlike them, the proposed differentiable regression forests perform soft data partition, so that an input-dependent partition function can be learned to handle heterogeneous data. In addition, the input feature space and the data abstractions at leaf nodes (local regressors) can be learned jointly, which ensures that the local input-output correlation is homogeneous at the leaf node.

Recently, end-to-end learning with CNN has become very popular and has been shown to be useful for improving the performance of various computer vision tasks, such as image classification~\cite{Ref:KrizhevskySH12}, semantic segmentation~\cite{Ref:LongSD15} and object detection~\cite{Ref:RenHG017,Ref:DaiLHS16}. Our differentiable regression forests can be seamlessly integrated with any deep networks, which enables us to conduct an end-to-end deep age estimation model, which we name Deep Regression Forests (DRFs). To build such a tree based model, we apply an alternating optimization strategy: first we fix the leaf nodes and optimize the data partitions at split nodes as well as the CNN parameters (feature learning) by Back-propagation; Then, we fix the split nodes and optimize the data abstractions at leaf nodes (local regressors) by Variational Bounding~\cite{Ref:Jordan99,Ref:Yuille03}. These two learning steps are alternatively performed to jointly optimize feature learning and regression modeling for age estimation.

We evaluate our algorithm on three standard benchmarks for real age estimation methods: MORPH~\cite{Ref:MORPH06}, FG-NET~\cite{Ref:PanisLTC16} and the Cross-Age Celebrity Dataset (CACD)~\cite{Ref:ChenCH15}. Experimental results demonstrate that our algorithm outperforms several state-of-the-art methods on these three benchmarks.

Our algorithm was inspired by Deep Neural Decision Forests (dNDFs)~\cite{Ref:Kontschieder15} and Label Distribution Learning Forests (LDLFs)~\cite{Ref:Shen17}, but differs in its objective (regression \emph{vs} classification/label distribution learning). Extending differentiable decision trees to deal with regression is non-trivial, as the distribution of the output space for regression is continuous, but the distribution of the output space for the two classification tasks is discrete. The contribution of this paper is three folds:

1) We propose Deep Regression Forests (DRFs), an end-to-end model, to deal with heterogeneous data by jointly learning input-dependant data partition at split nodes and data abstraction at leaf nodes.

2) Based on Variational Bounding, the convergence of our update rule for leaf nodes in DRFs is mathematically guaranteed.

3) We apply DRFs on three standard age estimation benchmarks, and achieve state-of-the-art results.

\section{Related Work}
\textbf{Age Estimation}
One way to tackle precise facial age estimation is to search for a kernel-based global non-linear mapping, like kernel support vector regression \cite{Ref:GuoMFH09} or kernel partial least squares regression \cite{Ref:GuoM11}. The basic idea is to learn a low-dimensional embedding of the aging manifold \cite{Ref:Guo08}. However, global non-linear mapping algorithms may be biased \cite{Ref:Huang17}, due to the heterogenous properties of the input data. Another way is adopting divide-and-conquer approaches, which partition the data space and learn multiple local regressors. But hierarchical regression~\cite{Ref:HanOLJ15} or tree based regression~\cite{Ref:Montillo09} approaches made hard partitions according to ages, which is problematic because the subsets of facial images may not be homogeneous for learning local regressors. Huang \emph{et al.}~\cite{Ref:Huang17} proposed Soft-margin Mixture of Regressions (SMMR) to address this issue, which found homogenous partitions in the joint input-output space, and learned a local regressor for each partition. But their regression model cannot be integrated with any deep networks as an end-to-end model.

Several researchers formulated age estimation as an ordinal regression problem \cite{Ref:Chang11,Ref:Niu16,Ref:ChenZDLR17}, because the relative order among the age labels is also important information. They trained a series of binary classifiers to partition the samples according to ages, and estimated ages by summing over the classifier outputs. Thus, ordinal regression is limited by its lack of scalability~\cite{Ref:Huang17}. Some other researchers formulated age estimation as a label distribution learning (LDL) problem~\cite{Ref:geng2016label}, which paid attention to modeling the cross-age correlations, based on the observation that faces at close ages look similar. LDL based age estimation methods~\cite{Ref:Geng2013Facial,Ref:Geng13,Ref:Yang16,Ref:Shen17} achieved promising results, but the label distribution model is usually inflexible in adapting to complex face data domains with diverse cross-age correlations~\cite{Ref:He17}.

With the rapid development of deep networks, more and more end-to-end CNN based age estimation methods~\cite{Ref:Rothe16,Ref:Niu16,Ref:Agustsson17} have been proposed to address this non-linear regression problem. But how to deal with heterogeneous data is still an open issue.

\textbf{Random Forests}
Random forests are an ensemble of randomized decision trees~\cite{Ref:Breiman01}. Each decision tree consists of several split nodes and leaf nodes. Tree growing is usually based on greedy algorithms which make locally-optimal hard data partition decisions at each split node. Thus, this makes it intractable to integrate decision trees and deep networks in an end-to-end learning manner. The newly proposed Deep Neural Decision Forests (dNDFs)~\cite{Ref:Kontschieder15} addressed this shortcoming by defining a soft partition function at each split node, which enabled the decision trees to be differentiable, allowing joint learning with deep networks. Shen \emph{et al.}~\cite{Ref:Shen17} then extended this differentiable decision tree to address label distribution learning problems. As we mentioned in Sec.~\ref{sec:intro}, our DRFs model is inspired by these two works, but differs in the objective (regression \emph{vs} classification/label distribution learning). One recent work proposed Neural Regression Forest (NRF)~\cite{Ref:Roy16} for depth estimation, which is similar to our DRFs. But mathematically, the convergence of their update rule for leaf nodes was not guaranteed.
\section{Deep Regression Forests}
In this section, we first introduce how to learn a single differentiable regression tree, then describe how to learn tree ensembles to form a forest.
\subsection{Problem Formulation}\label{sec:pro_form}
\begin{figure}[!t]
\centering
\includegraphics[trim=4cm 2cm 4cm 8cm, clip=true, width=1.0\linewidth]{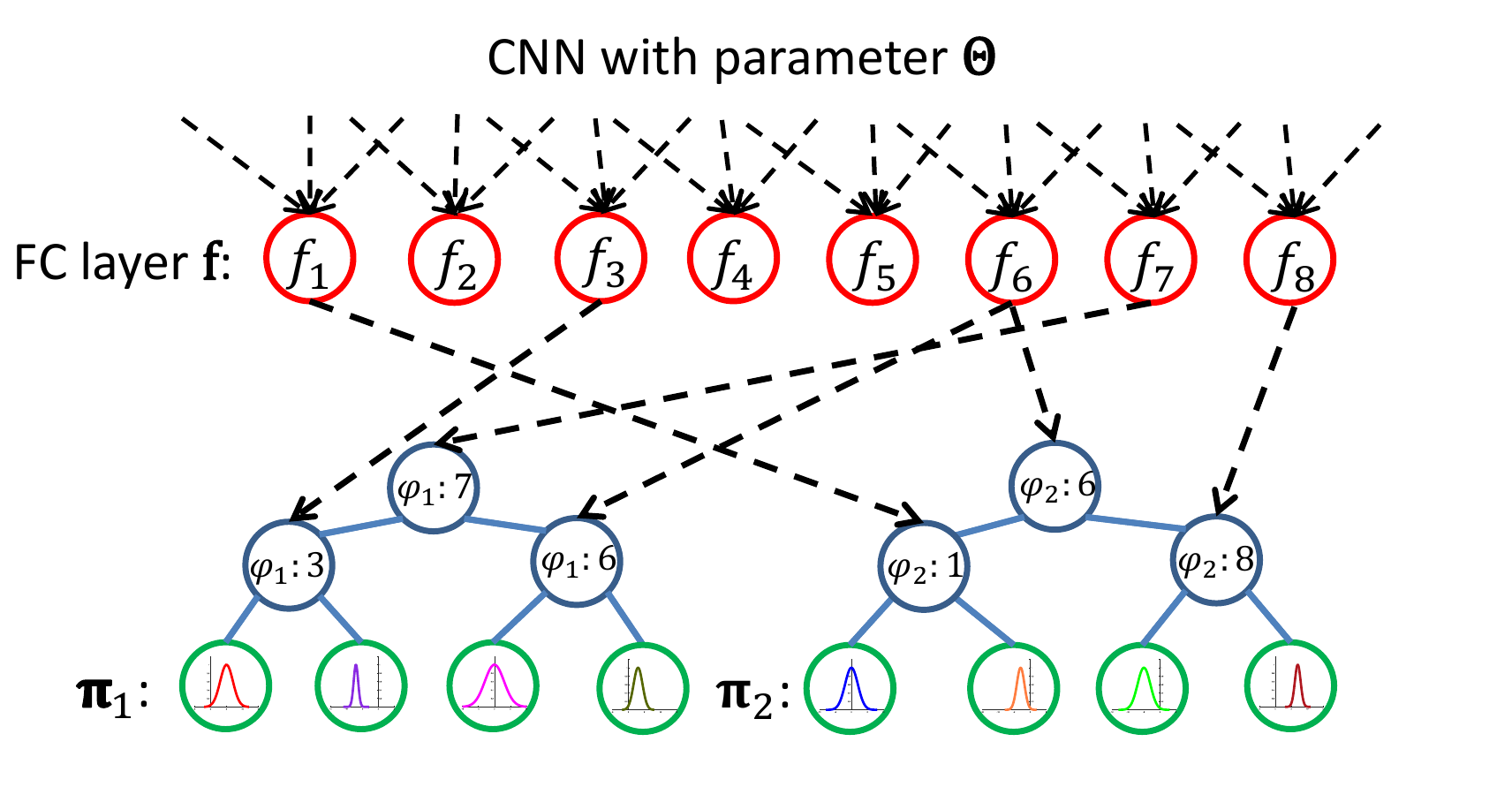}
\caption{Illustration of a deep regression forest (DRF). The top red circles denote the output units of the function $\mathbf{f}$ parameterized by $\bm{\Theta}$. Here, they are the units of a fully-connected (FC) layer in a CNN. The blue and green circles are split nodes and leaf nodes, respectively. Two index functions $\varphi_1$ and $\varphi_2$ are assigned to these two trees respectively. The black dash arrows indicate the correspondence between the split nodes of these two trees and the output units of the FC layer. Note that, one output unit may correspond to the split nodes belonging to different trees. Each tree has independent leaf node distribution $\bm{\pi}$ (denoted by distribution curves in leaf nodes). The output of the forest is a mixture of the tree predictions. $\mathbf{f}(\cdot;\bm{\Theta})$ and $\bm{\pi}$ are learned jointly in an end-to-end manner.}\label{fig:DRForest}
\end{figure}

\begin{figure}[!t]
\centering
\includegraphics[trim=5cm 4cm 5cm 7cm, clip=true, width=1.0\linewidth]{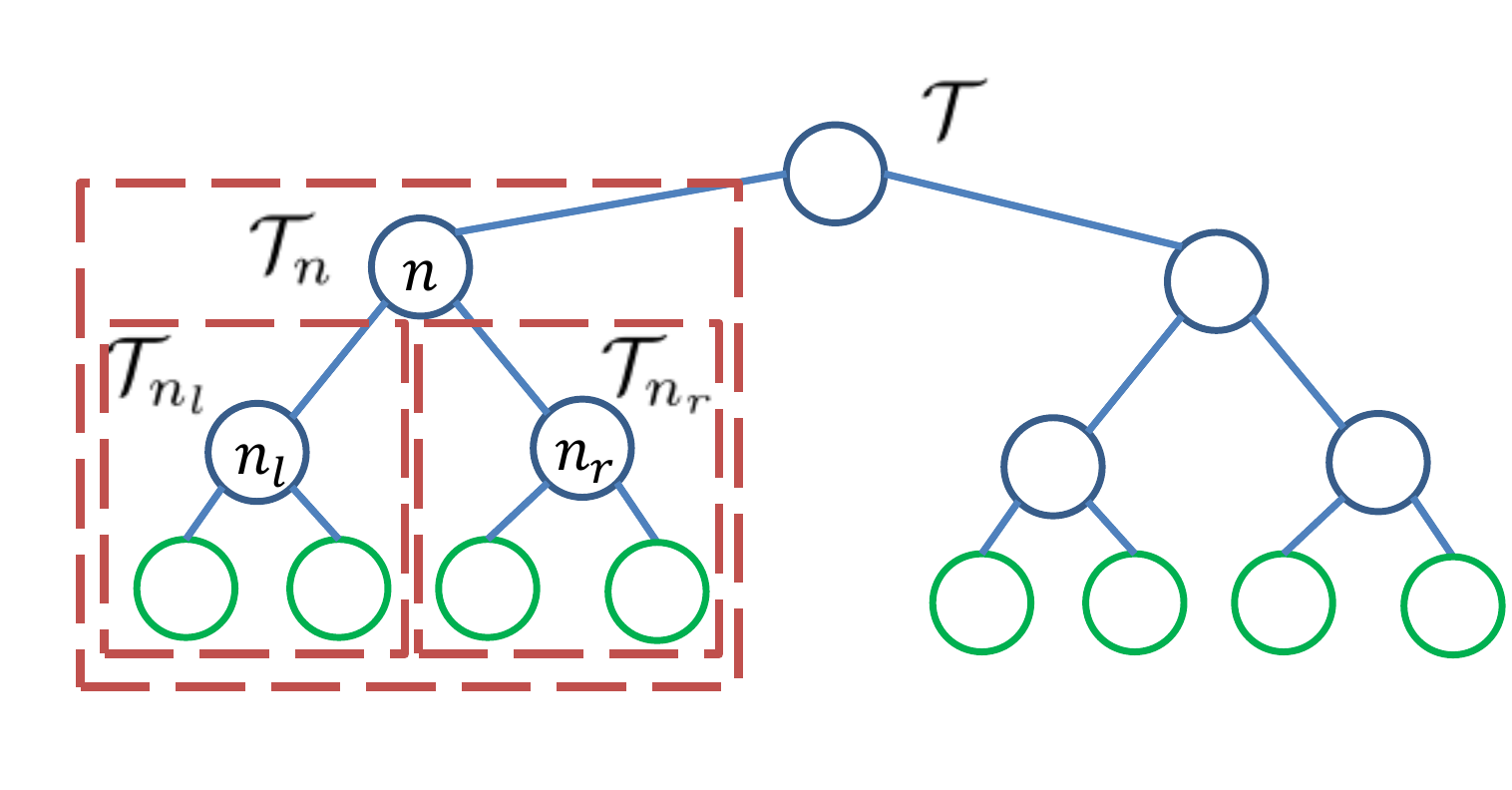}
\caption{The subtree rooted at node $n$: $\mathcal {T}_n$ and its left and right subtrees: $\mathcal {T}_{n_l}$ and $\mathcal {T}_{n_r}$.}\label{fig:Subtree}
\end{figure}

Let $\mathcal {X} = \mathbb{R}^{d_x}$ and $\mathcal {Y} = \mathbb{R}^{d_y}$ denote the input and output spaces, respectively. We consider a regression problem, where for each input sample $\mathbf{x}\in\mathcal {X}$, there is an output target $\mathbf{y}\in\mathcal {Y}$. The objective of regression is to find a mapping function $\mathbf{g}:\mathbf{x}\rightarrow\mathbf{y}$ between an input sample $\mathbf{x}$ and its output target $\mathbf{y}$. A standard way to address this problem is to model the conditional probability function $p(\mathbf{y}|\mathbf{x})$, so that the mapping is given by
\begin{equation}
\hat{\mathbf{y}} = \mathbf{g}(\mathbf{x}) = \int\mathbf{y}p(\mathbf{y}|\mathbf{x})d\mathbf{y}.
\end{equation}
We propose to model this conditional probability by a decision tree based structure $\mathcal {T}$. A decision regression tree consists of a set of split nodes $\mathcal {N}$ and a set of leaf nodes $\mathcal {L}$. Each split node $n\in\mathcal {N}$ defines a split function $s_n(\cdot;\bm{\Theta}):\mathcal {X}\rightarrow[0,1]$ parameterized by $\bm{\Theta}$ to determine whether a sample is sent to the left or right subtree. Each leaf node $\ell\in\mathcal {L}$ contains a probability density distribution $\pi_{\ell}(\mathbf{y})$ over $\mathcal {Y}$, i.e, $\int\pi_{\ell}(\mathbf{y})d\mathbf{y}=1$. Following~\cite{Ref:Kontschieder15,Ref:Shen17}, we use a soft split function  $s_n(\mathbf{x};\bm{\Theta}) = \sigma(f_{\varphi(n)}(\mathbf{x};\bm{\Theta}))$, where $\sigma(\cdot)$ is a sigmoid function, $\varphi(\cdot)$ is an index function to bring the $\varphi(n)$-th output of function $\mathbf{f}(\mathbf{x};\bm{\Theta})$ in correspondence with a split node $n$, and $\mathbf{f}:\mathbf{x}\rightarrow\mathbb{R}^M$ is a real-valued feature learning function depending on the sample $\mathbf{x}$ and the parameter $\bm{\Theta}$. $\mathbf{f}$ can take any forms. In our DRFs, it is a CNN and $\bm{\Theta}$ is the network parameter. The index function $\varphi(\cdot)$ specifies the correspondence between the split nodes and output units of $\mathbf{f}$ (it is initialized randomly before tree learning). An example to demonstrate the sketch chart of our DRFs as well as $\varphi(\cdot)$ is shown in Fig.~\ref{fig:DRForest} (There are two trees with index functions, $\varphi_1$ and $\varphi_2$ for each). Then, the probability of the sample $\mathbf{x}$ falling into leaf node $\ell$ is given by
\begin{equation}
P(\ell|\mathbf{x};\bm{\Theta})=\prod_{n\in\mathcal {N}}s_n(\mathbf{x};\bm{\Theta})^{\mathbf{1}(\ell\in\mathcal {L}_{n_l})}(1-s_n(\mathbf{x};\bm{\Theta}))^{\mathbf{1}(\ell\in\mathcal {L}_{n_r})},
\end{equation}
where $\mathbf{1}(\cdot)$ is an indicator function and $\mathcal {L}_{n_l}$ and $\mathcal {L}_{n_r}$ denote the sets of leaf nodes held by the subtrees $\mathcal {T}_{n_l}$, $\mathcal {T}_{n_r}$ rooted at the left and right children $n_l, n_r$ of node $n$ (shown in Fig.~\ref{fig:Subtree}), respectively. The conditional probability function $p(\mathbf{y}|\mathbf{x};\mathcal {T})$ given by the tree $\mathcal {T}$ is
\begin{equation}
p(\mathbf{y}|\mathbf{x};\mathcal {T})=\sum_{\ell\in\mathcal{L}}P(\ell|\mathbf{x};\bm{\Theta})\pi_{\ell}(\mathbf{y}).
\end{equation}
Then the mapping between $\mathbf{x}$ and $\mathbf{y}$ modeled by tree $\mathcal {T}$ is given by
\begin{equation}
\hat{\mathbf{y}} = \mathbf{g}(\mathbf{x};\mathcal {T}) = \int\mathbf{y}p(\mathbf{y}|\mathbf{x};\mathcal {T})d\mathbf{y}.
\end{equation}

\subsection{Tree Optimization}\label{sec:tree_opt}
Given a training set $\mathcal {S}=\{(\mathbf{x}_i,\mathbf{y}_i)\}_{i=1}^N$, learning a regression tree $\mathcal {T}$ described in Sec.~\ref{sec:pro_form} leads to minimizing the following negative log likelihood loss:
\begin{eqnarray}\label{eqn:tree_loss}
R(\bm{\pi},\bm{\Theta};\mathcal{S})=-\frac{1}{N}\sum_{i=1}^{N}\log(p(\mathbf{y}_{i}|\mathbf{x}_{i}, \mathcal {T}))\nonumber\\
=-\frac{1}{N}\sum_{i=1}^{N}\log\big(\sum_{\ell\in\mathcal{L}}P(\ell|\mathbf{x}_i;\bm{\Theta})\pi_{\ell}(\mathbf{y}_i)\big),
\end{eqnarray}
where $\bm{\pi}$ denotes the density distributions contained by all the leaf nodes $\mathcal {L}$. Note that, optimizing $R(\bm{\pi},\bm{\Theta};\mathcal{S})$ requires estimating both the split node parameter $\bm{\Theta}$ and the density distributions $\bm{\pi}$ held by leaf nodes, i.e.,
\begin{equation}\label{eqn:best_para}
(\bm{\Theta}^{\ast},\bm{\pi}^{\ast}) = \arg\min_{\bm{\Theta},\bm{\pi}}R(\bm{\pi},\bm{\Theta};\mathcal{S}).
\end{equation}
To solve Eqn.~\ref{eqn:best_para}, we alternate the following two steps: (1) fixing $\bm{\pi}$ and optimizing $\bm{\Theta}$; (2) fixing $\bm{\Theta}$ and optimizing $\bm{\pi}$, until convergence or a maximum number of iterations is reached.

\subsubsection{Learning Split Nodes by Gradient Descent}\label{sec:split_node}
Now, we discuss how to learn the parameter $\bm{\Theta}$ for split nodes, when the density distributions held by the leaf nodes $\bm{\pi}$ are fixed. Thanks to the soft split function, the tree loss $R(\bm{\pi},\bm{\Theta};\mathcal{S})$ is differentiable with respect to $\bm{\Theta}$. The gradient of the loss is computed by the chain rules as follows:
\begin{equation}\label{eqn:chain_rule}
\frac{\partial{R}(\bm{\pi},\bm{\Theta};\mathcal {S})}{\partial{\bm{\Theta}}}=\sum_{i=1}^N\sum_{n\in\mathcal {N}}\frac{\partial{R}(\bm{\pi},\bm{\Theta};\mathcal {S})}{{\partial}f_{\varphi(n)}(\mathbf{x}_i;\bm{\Theta})}\frac{{\partial}f_{\varphi(n)}(\mathbf{x}_i;\bm{\Theta})}{\partial{\bm{\Theta}}}.
\end{equation}
Note that in the right part of Eqn.~\ref{eqn:chain_rule}, only the first term depends on the tree and the second term depends only on the specific form of the function $f_{\varphi(n)}$. The first term is computed by
%\begin{eqnarray}\label{eqn:gradient}
%\frac{\partial{R}(\bm{\pi},\bm{\Theta};\mathcal {S})}{{\partial}f_{\varphi(n)}(\mathbf{x}_i;\bm{\Theta})}=\frac{1}{N}\Big(s_n(\mathbf{x}_i;\bm{\Theta})\frac{p(\mathbf{y}_i|\mathbf{x}_i;\mathcal {T}_n^r)}{p(\mathbf{y}_i|\mathbf{x}_i;\mathcal {T})}\nonumber\\
%-\big(1-s_n(\mathbf{x}_i;\bm{\Theta})\big)\frac{p(\mathbf{y}_i|\mathbf{x}_i;\mathcal {T}_n^l)}{p(\mathbf{y}_i|\mathbf{x}_i;\mathcal {T})}\Big),
%\end{eqnarray}
\begin{equation}\label{eqn:gradient}
\frac{\partial{R}(\bm{\pi},\bm{\Theta};\mathcal {S})}{{\partial}f_{\varphi(n)}(\mathbf{x}_i;\bm{\Theta})}=\frac{1}{N}\Big(s_n(\mathbf{x}_i;\bm{\Theta})\Gamma^i_{n_r} - \big(1-s_n(\mathbf{x}_i;\bm{\Theta})\big)\Gamma^i_{n_l}\Big),
\end{equation}
where for a generic node $n\in\mathcal{N}$
\begin{equation}
\Gamma^i_{n}=\frac{p(\mathbf{y}_i|\mathbf{x}_i;\mathcal {T}_n)}{p(\mathbf{y}_i|\mathbf{x}_i;\mathcal {T})}=\frac{\sum_{\ell\in\mathcal{L}_n}P(\ell|\mathbf{x}_i;\bm{\Theta})\pi_{\ell}(\mathbf{y}_i)}{p(\mathbf{y}_i|\mathbf{x}_i;\mathcal {T})}.
\end{equation}
$\Gamma^i_{n}$ can be efficiently computed for all nodes $n$ in the tree $\mathcal {T}$ by a single pass over the tree. Observing that $\Gamma^i_{n}=\Gamma^i_{n_l} + \Gamma^i_{n_r}$, the computation for $\Gamma^i_{n}$ can be started at the leaf nodes and conducted in a bottom-up manner. Based on Eqn.~\ref{eqn:gradient}, the split node parameters $\bm{\Theta}$ can be learned by standard Back-propagation.
\subsubsection{Learning Leaf Nodes by Variational Bounding}
By fixing the split node parameters $\bm{\Theta}$, Eqn.~\ref{eqn:best_para} becomes a constrained optimization problem:
\begin{equation}\label{eqn:best_pi}
\min_{\bm{\pi}}R(\bm{\pi},\bm{\Theta};\mathcal {S}),\textbf{s.t.}, \forall\ell,\int\pi_{\ell}(\mathbf{y})d\mathbf{y}=1.
\end{equation}
For efficient computation, we represent each density distribution $\pi_{\ell}(\mathbf{y})$ by a parametric model. Since ideally each leaf node corresponds to a compact homogeneous subset, we assume that the density distribution $\pi_{\ell}(\mathbf{y})$
in each leaf node is a Gaussian distribution, i.e.,
\begin{equation}
\pi_{\ell}(\mathbf{y})=\frac{1}{\sqrt{(2\pi)^{k}\textrm{det}(\bm{\Sigma}_{\ell})}}\exp(-\frac{1}{2}(\mathbf{y-\bm{\mu}_{\ell}})^{\mathrm{T}}\Sigma_{\ell}^{-1}(\mathbf{y-\bm{\mu}_{\ell}})),
\end{equation}
where $\bm{\mu}_{\ell}$ and $\bm{\Sigma}_{\ell}$ are the mean and the covariance matrix of the Gaussian distribution. Based on this assumption, Eqn.~\ref{eqn:best_pi} is equivalent to minimizing $R(\bm{\pi},\bm{\Theta};\mathcal {S})$ w.r.t. $\bm{\mu}_{\ell}$ and $\bm{\Sigma}_{\ell}$. Now, we propose to address this optimization problem by Variational Bounding~\cite{Ref:Jordan99,Ref:Yuille03}. In variational
bounding, an original objective function to be minimized gets replaced by a sequence of bounds minimized in an iterative manner. To obtain an upper bound of $R(\bm{\pi},\bm{\Theta};\mathcal {S})$, we apply Jensen's inequality to it:
\begin{align}\label{eqn:jensen}
R(\bm{\pi},\bm{\Theta};\mathcal{S})=-\frac{1}{N}\sum_{i=1}^{N}\log\big(\sum_{\ell\in\mathcal{L}}P(\ell|\mathbf{x}_i;\bm{\Theta})\pi_{\ell}(\mathbf{y}_i)\big)\nonumber\\
=-\frac{1}{N}\sum_{i=1}^{N}\log\Big(\sum_{\ell\in\mathcal{L}}\zeta_{\ell}(\bar{\bm{\pi}};\mathbf{x}_i,\mathbf{y}_i)\frac{P(\ell|\mathbf{x}_i;\bm{\Theta})\pi_{\ell}(\mathbf{y}_i)}{\zeta_{\ell}(\bar{\bm{\pi}};\mathbf{x}_i,\mathbf{y}_i)}\Big)\nonumber\\
\leq  -\frac{1}{N}\sum_{i=1}^{N}\sum_{\ell\in\mathcal{L}}\zeta_{\ell}(\bar{\bm{\pi}};\mathbf{x}_i,\mathbf{y}_i)\log\Big(\frac{P(\ell|\mathbf{x}_i;\bm{\Theta})\pi_{\ell}(\mathbf{y}_i)}{\zeta_{\ell}(\bar{\bm{\pi}};\mathbf{x}_i,\mathbf{y}_i)}\Big)\nonumber\\
=R(\bar{\bm{\pi}},\bm{\Theta};\mathcal{S})-\frac{1}{N}\sum_{i=1}^{N}\sum_{\ell\in\mathcal{L}}\zeta_{\ell}(\bar{\bm{\pi}};\mathbf{x}_i,\mathbf{y}_i)\log\Big(\frac{\pi_{\ell}(\mathbf{y}_i)}{\bar{\pi}_{\ell}(\mathbf{y}_i)} \Big),
\end{align}
where $\zeta_{\ell}(\bm{\pi};\mathbf{x}_i,\mathbf{y}_i) = \frac{P(\ell|\mathbf{x}_i;\bm{\Theta})\pi_{\ell}(\mathbf{y}_i)}{p(\mathbf{y}_i|\mathbf{x}_i;\mathcal {T})}$. Note that $\zeta_{\ell}(\bm{\pi};\mathbf{x}_i,\mathbf{y}_i)$ has the property that $\zeta_{\ell}(\bm{\pi};\mathbf{x}_i,\mathbf{y}_i) \in [0, 1]$ and $\sum_{\ell\in\mathcal{L}}\zeta_{\ell}(\bm{\pi};\mathbf{x}_i,\mathbf{y}_i)=1$ to ensure that Eqn.~\ref{eqn:jensen} holds Jensen's inequality.
Let us define
\begin{equation}
\phi(\bm{\pi},\bar{\bm{\pi}}) = R(\bar{\bm{\pi}},\bm{\Theta};\mathcal{S})-\frac{1}{N}\sum_{i=1}^{N}\sum_{\ell\in\mathcal{L}}\zeta_{\ell}(\bar{\bm{\pi}};\mathbf{x}_i,\mathbf{y}_i)\log\Big(\frac{\pi_{\ell}(\mathbf{y}_i)}{\bar{\pi}_{\ell}(\mathbf{y}_i)} \Big).
\end{equation}
Then $\phi(\bm{\pi},\bar{\bm{\pi}})$ is an upper bound for $R(\bm{\pi},\bm{\Theta};\mathcal {S})$, which has the properties that for any $\bm{\pi}$ and $\bar{\bm{\pi}}$, $\phi(\bm{\pi},\bar{\bm{\pi}})\geq \phi(\bm{\pi},\bm{\pi})=R(\bm{\pi},\bm{\Theta};\mathcal {S})$ and $\phi(\bar{\bm{\pi}},\bar{\bm{\pi}})= R(\bar{\bm{\pi}},\bm{\Theta};\mathcal {S})$. These two properties hold the conditions for Variational Bounding.

Recall that we parameterize $\pi_{\ell}(\mathbf{y})$ by two parameters: the mean $\bm{\mu}_{\ell}$ and the covariance matrix $\bm{\Sigma}_{\ell}$. Let $\bm{\mu}$ and $\bm{\Sigma}$ denote these two parameters held by all the leaf nodes $\mathcal{L}$. We define $\psi(\bm{\mu},\bar{\bm{\mu}}) = \phi(\bm{\pi},\bar{\bm{\pi}})$, then $\psi(\bm{\mu},\bar{\bm{\mu}}) \geq \phi(\bm{\pi},\bm{\pi}) = \psi(\bm{\mu},\bm{\mu}) = R(\bm{\pi},\bm{\Theta};\mathcal {S})$, which indicates that $\psi(\bm{\mu},\bar{\bm{\mu}})$ is also an upper bound for $R(\bm{\pi},\bm{\Theta};\mathcal {S})$. Assume that we are at a point $\bm{\mu}^{(t)}$ corresponding to the $t$-th iteration, then $\psi(\bm{\mu},\bm{\mu}^{(t)})$ is an upper bound for $R(\bm{\mu},\bm{\Theta};\mathcal {S})$. In the next iteration, $\bm{\mu}^{(t+1)}$ is chosen such that $\psi(\bm{\mu}^{(t+1)},\bm{\mu})\leq R(\bm{\mu}^{(t)},\bm{\Theta};\mathcal {S})$, which implies $R(\bm{\mu}^{(t+1)},\bm{\Theta};\mathcal {S})\leq R(\bm{\mu}^{(t)},\bm{\Theta};\mathcal {S})$. Therefore, we can minimize $\psi(\bm{\mu},\bar{\bm{\mu}})$ instead of $R(\bm{\mu},\bm{\Theta};\mathcal {S})$ after ensuring that
$R(\bm{\mu}^{(t)},\bm{\Theta};\mathcal {S})=\psi(\bm{\mu}^{(t)},\bar{\bm{\mu}})$, i.e., $\bar{\bm{\mu}} = \bm{\mu}^{(t)}$. Thus, we have
\begin{equation}
\bm{\mu}^{(t+1)} = \arg\min_{\bm{\mu}}\psi(\bm{\mu}, \bm{\mu}^{(t)}).
\end{equation}
The partial derivative of $\psi(\bm{\mu}, \bm{\mu}^{(t)})$ w.r.t. $\bm{\mu}_{\ell}$ is computed by
\begin{align}
&\frac{\partial\psi(\bm{\mu}, \bm{\mu}^{(t)})}{\partial\bm{\mu}_{\ell}} = \frac{\partial\phi(\bm{\pi}, \bm{\pi}^{(t)})}{\partial\bm{\mu}_{\ell}}\nonumber\\
&=-\frac{1}{N}\sum_{i=1}^{N}\zeta_{\ell}(\bm{\pi}^{(t)};\mathbf{x}_i,\mathbf{y}_i)\frac{\partial\log(\pi_{\ell}(\mathbf{y}_i))}{\partial\bm{\mu}_{\ell}}\nonumber\\
&=-\frac{1}{N}\sum_{i=1}^{N}\zeta_{\ell}(\bm{\pi}^{(t)};\mathbf{x}_i,\mathbf{y}_i)\bm{\Sigma}^{-1}_{\ell}(\mathbf{y}_i - \bm{\mu}_{\ell}).
\end{align}
By setting $\frac{\partial\psi(\bm{\mu}, \bm{\mu}^{(t)})}{\partial\bm{\mu}_{\ell}} = \bm{0}$, where $\bm{0}$ denotes zero vector or matrix, we have
\begin{equation}\label{eqn:mean}
\bm{\mu}_{\ell}^{(t+1)}=\frac{\sum_{i=1}^N\zeta_{\ell}(\bm{\pi}^{(t)};\mathbf{x}_i,\mathbf{y}_i)\mathbf{y}_i}{\sum_{i=1}^N\zeta_{\ell}(\bm{\pi}^{(t)};\mathbf{x}_i,\mathbf{y}_i)}.
\end{equation}
Similarly, we define $\xi(\bm{\Sigma}, \bar{\bm{\Sigma}})=\phi(\bm{\pi},\bar{\bm{\pi}})$, then
\begin{equation}
\bm{\Sigma}^{(t+1)} = \arg\min_{\bm{\Sigma}}\xi(\bm{\Sigma}, \bm{\Sigma}^{(t)}).
\end{equation}
The partial derivative of $\xi(\bm{\Sigma}, \bm{\Sigma}^{(t)})$ w.r.t. $\bm{\Sigma}_{\ell}$ is obtained by
\begin{align}
&\frac{\partial\xi(\bm{\Sigma}, \bm{\Sigma}^{(t)})}{\partial\bm{\Sigma}_{\ell}} = \frac{\partial\phi(\bm{\pi}, \bm{\pi}^{(t)})}{\partial\bm{\Sigma}_{\ell}}\nonumber\\
&=-\frac{1}{N}\sum_{i=1}^{N}\zeta_{\ell}(\bm{\pi}^{(t)};\mathbf{x}_i,\mathbf{y}_i)\frac{\partial\log(\pi_{\ell}(\mathbf{y}_i))}{\partial\bm{\Sigma}_{\ell}}\nonumber\\
&=-\frac{1}{N}\sum_{i=1}^{N}\zeta_{\ell}(\bm{\pi}^{(t)};\mathbf{x}_i,\mathbf{y}_i)\big[-\frac{1}{2}\bm{\Sigma}_{\ell}^{-1}\nonumber\\
&+\frac{1}{2}\bm{\Sigma}_{\ell}^{-1}(\mathbf{y}_i-\bm{\mu}_{\ell}^{(t+1)})(\mathbf{y}_i-\bm{\mu}_{\ell}^{(t+1)})^{\mathrm{T}}\bm{\Sigma}_{\ell}^{-1}\big]
\end{align}
By Setting $\frac{\partial\xi(\bm{\Sigma}, \bm{\Sigma}^{(t)})}{\partial\bm{\Sigma}_{\ell}} = \bm{0}$, we have
\begin{equation}
\sum_{i=1}^{N}\zeta_{\ell}(\bm{\pi}^{(t)};\mathbf{x}_i,\mathbf{y}_i)\big[-\bm{\Sigma}_{\ell}+(\mathbf{y}_i-\bm{\mu}_{\ell}^{(t+1)})(\mathbf{y}_i-\bm{\mu}_{\ell}^{(t+1)})^{\mathrm{T}} \big] = \bm{0},
\end{equation}
which leads to
\begin{equation} \label{eqn:covar}
\bm{\Sigma}_{\ell}^{(t+1)}=\frac{\sum_{i=1}^{N}\zeta_{\ell}(\bm{\pi}^{(t)};\mathbf{x}_i,\mathbf{y}_i)(\mathbf{y}_i-\bm{\mu}_{\ell}^{(t+1)})(\mathbf{y}_i-\bm{\mu}_{\ell}^{(t+1)})^{\mathrm{T}}}{\sum_{i=1}^{N}\zeta_{\ell}(\bm{\pi}^{(t)};\mathbf{x}_i,\mathbf{y}_i)}.
\end{equation}
Eqn.~\ref{eqn:mean} and Eqn.~\ref{eqn:covar} are the update rule for the density distribution $\bm{\pi}$ held by all leaf nodes, which are step-size free and fast-converged. One issue remained is how to initialize the starting point $\bm{\mu}_{\ell}^{(0)}$ and $\bm{\Sigma}_{\ell}^{(0)}$. The simplest way is to do k-means clustering on $\{\mathbf{y}_i\}_{i=1}^N$ to obtain $|\mathcal{L}|$ subsets, then initialize $\bm{\mu}_{\ell}^{(0)}$ and $\bm{\Sigma}_{\ell}^{(0)}$ according to cluster assignment, i.e., let $\mathbb{I}_i$ denote cluster index assigned to $\mathbf{y}_i$, then
\begin{equation} \label{eqn:init_pi}
\begin{split}
&\bm{\mu}_{\ell}^{(0)} = \frac{\sum_{i=1}^N\mathbf{1}(\mathbb{I}_i=\ell)\mathbf{y}_i}{\sum_{i=1}^N\mathbf{1}(\mathbb{I}_i=\ell)},\\
&\bm{\Sigma}_{\ell}^{(0)}=\frac{\sum_{i=1}^N\mathbf{1}(\mathbb{I}_i=\ell)(\mathbf{y}_i-\bm{\mu}_{\ell}^{(0)})(\mathbf{y}_i-\bm{\mu}_{\ell}^{(0)})^{\mathrm{T}}}{\sum_{i=1}^N\mathbf{1}(\mathbb{I}_i=\ell)}.\\
\end{split}
\end{equation}
This initialization can be understood in this way that we first perform data partition only according to ages by k-means, and then the input facial feature space and output age space are jointly learned to find homogeneous partitions during tree building.

\subsubsection{Learning a Regression Forest}
A regression forest is an ensemble of regression trees $\mathcal {F}=\{\mathcal {T}^1,\ldots,\mathcal {T}^K\}$, where all trees can possibly share the same parameters in $\bm{\Theta}$, but each tree can have a different set of split functions (assigned by $\varphi$, as shown in Fig.~\ref{fig:DRForest}), and independent leaf node distribution $\bm{\pi}$. We define the loss function for a forest as the averaged loss functions of all individual trees: $R_{\mathcal {F}}=\frac{1}{K}\sum_{k=1}^KR_{\mathcal{T}^k}$,
%\begin{equation}
%R_{\mathcal {F}}=\frac{1}{K}\sum_{k=1}^KR_{\mathcal{T}_k},
%\end{equation}
where $R_{\mathcal{T}^k}$ is the loss function for tree $\mathcal{T}^k$ defined by Eqn.~\ref{eqn:tree_loss}. Learning the forest $\mathcal {F}$ also follows the alternating optimization strategy described in Sec.~\ref{sec:tree_opt}.

\begin{algorithm}
\caption{The training procedure of a DRF.}
\label{fig:algorithm}
\begin{algorithmic}
\REQUIRE $\mathcal{S}$: training set, $n_B$: the number of mini-batches to update $\bm{\pi}$
\STATE Initialize $\bm{\Theta}$ randomly and $\bm{\pi}$ by Eqn.~\ref{eqn:init_pi}. Set $\mathcal{B}=\{\emptyset\}$
\WHILE {Not converge}
\WHILE {$|\mathcal{B}|<n_B$}
\STATE Randomly select a mini-batch $B$ from $\mathcal{S}$
\STATE Update $\bm{\Theta}$ by computing gradient (Eqn.~\ref{eqn:gradient_forest}) on $B$
\STATE $\mathcal{B}=\mathcal{B}\bigcup{B}$
\ENDWHILE
\STATE Update $\bm{\pi}$ by iterating Eqn.~\ref{eqn:mean} and Eqn.~\ref{eqn:covar} on $\mathcal{B}$
\STATE $\mathcal{B}=\{\emptyset\}$
\ENDWHILE
\end{algorithmic}
\end{algorithm}

To learn $\bm{\Theta}$, by referring to Fig.~\ref{fig:DRForest} and our derivation in Sec.~\ref{sec:split_node}, we have
\begin{equation}\label{eqn:gradient_forest}
\frac{\partial{R}_{\mathcal {F}}}{\partial{\bm{\Theta}}}=\frac{1}{K}\sum_{i=1}^N\sum_{k=1}^K\sum_{n\in\mathcal {N}_k}\frac{\partial{R}_{\mathcal{T}^k}}{{\partial}f_{\varphi_k(n)}(\mathbf{x}_i;\bm{\Theta})}\frac{{\partial}f_{\varphi_k(n)}(\mathbf{x}_i;\bm{\Theta})}{\partial{\bm{\Theta}}},
\end{equation}
where $\mathcal {N}_k$ and $\varphi_k(\cdot)$ are the split node set and the index function of $\mathcal{T}^k$, respectively. The index function $\varphi_k(\cdot)$ for each tree is randomly assigned before tree learning, which means the split nodes of each tree are connected to a randomly selected subset of output units of $\mathbf{f}$. This strategy is similar to the random subspace method~\cite{Ref:Ho98}, which can increase the randomness in training to reduce the risk of overfitting.

As each tree in the forest $\mathcal {F}$ has its own leaf node distribution $\bm{\pi}$, we update them independently according to Eqn.~\ref{eqn:mean} and Eqn.~\ref{eqn:covar}.
In our implementation, we do not conduct this update scheme on the whole dataset $\mathcal{S}$ but on a set of mini-batches $\mathcal{B}$. The training procedure of a DRF is shown in Algorithm.~\ref{fig:algorithm}.

In the testing stage, the output of the forest $\mathcal {F}$ is given by averaging the predictions from all the individual trees:
\begin{align}
&\hat{\mathbf{y}} = \mathbf{g}(\mathbf{x};\mathcal {F})=\frac{1}{K}\sum_{k=1}^K\mathbf{g}(\mathbf{x};\mathcal {T}^k)\nonumber\\
&=\frac{1}{K}\sum_{k=1}^K\int\mathbf{y}p(\mathbf{y}|\mathbf{x};\mathcal {T}^k)d\mathbf{y}\nonumber\\
&=\frac{1}{K}\sum_{k=1}^K\int\mathbf{y}\sum_{\ell\in\mathcal{L}^k}P(\ell|\mathbf{x};\bm{\Theta})\pi_{{\ell}}(\mathbf{y})d\mathbf{y}\nonumber\\
&=\frac{1}{K}\sum_{k=1}^K\sum_{\ell\in\mathcal{L}^k}P(\ell|\mathbf{x};\bm{\Theta})\bm{\mu}_{\ell},
\end{align}
where $\mathcal{L}^k$ is the leaf node set of the $k$-th tree. Here, we take the fact that the expectation of the Gaussian distribution $\pi_{\ell}(\mathbf{y})$ is $\bm{\mu}_{\ell}$.
\section{Experiments}
In this section we introduce the implementation details
and report the performance of the proposed algorithm as well as the comparison to other competitors.

\subsection{Implementation Details}
Our realization of DRFs is based on the public available ``caffe''~\cite{Ref:Jia2014caffe} framework.
Following the recent deep learning based age estimation method~\cite{Ref:Rothe16}, we use the VGG-16 Net~\cite{Ref:Simonyan14} as the CNN part of the proposed DRFs.
%We use the
%The proposed \emph{regression forests} is appended to the back of VGG~\cite{Ref:Simonyan14} network as a regular loss function,
%therefore the overall architecture can be trained in a end-to-end manner.
%

\textbf{Parameters Setting} The model-related hyper-parameters (and the default values we used)
are:
number of trees (5), tree depth (6),
number of output units produced by the feature learning function (128),
iterations to update leaf-node predictions (20),
number of mini-batches used to update leaf node predictions (50).
The network training based hyper-parameters (and the values we used) are: initial learning rate (0.05), mini-batch size (16), maximal iterations (30k). We decrease the learning rate ($\times 0.5$) every 10k iterations.

%We compare the performance of DRFs with other methods on three open facial age estimation datasets: MORPH~\cite{0.05}, FG-NET~\cite{xx} and CACD~\cite{xx}.
%Quantitative performance of facial age estimation is measured in terms of Mean-Average-Error (MAE) as well as Cumulative Score (CS).

\textbf{Preprocessing and Data Augmentation} Following the previous method~\cite{Ref:Niu16}, faces are firstly detected by
using a standard face detectior~\cite{Ref:ViolaJ01} and facial landmarks are localized by AAM~\cite{Ref:CootesET98}.
We perform face alignment to guarantee all eyeballs stay at the same position in the image.

Data augmentation is crucial to train good deep networks.
We augment the training data by:
(a) cropping images at random offsets,
(b) adding gaussian noise to the original images,
(c) randomly flipping (left-right).

%\textbf{The Unbalancing of Ages} The ages in original datasets are unbalanced.
%Take FG-NET as an example, almost 70\% of the samples are within $0~\sim19$ years old,
%less than 0.5\% are beyond 60.
%This unbalancing in labels will seriously influence the  performance of trained model.
%To offset the label unbalancing, we augment training data with age-associated augmentation factor
%$f_a(I) = \frac{\alpha}{percentage(I)}$
%Where $percentage(I)$ is percentage of sample image $I$'age in the whole dataset,
%$\alpha$ is a factor that controls the augmentation factor of the overall dataset.

\subsection{Experimental Results}
\subsubsection{Evaluation Metric}
The performance of age estimation is evaluated in terms of mean absolute error (MAE) as well
as Cumulative Score (CS).
MAE is the average absolute error over the testing set,
and the Cumulative Score is calculated by $\text{CS}(l) = \frac{K_l}{K} \cdot 100\%$, where $K$ is the total number of testing images and $K_l$ is the number of testing facial images whose absolute error between the estimated age and the ground truth age is not greater than $l$ years. Here, we set the same error level 5 as in~\cite{Ref:Chang11,Ref:Chen2013Cumulative,Ref:huerta2014facial}, i.e., $l=5$. Note that, because only some methods reported the Cumulative Score, we are only able to give CS values for some competitors.

\subsubsection{Performance Comparison}
\begin{figure}[!t]
\centering
\includegraphics[trim=3cm 5cm 4cm 5cm, clip=true, width=1.0\linewidth]{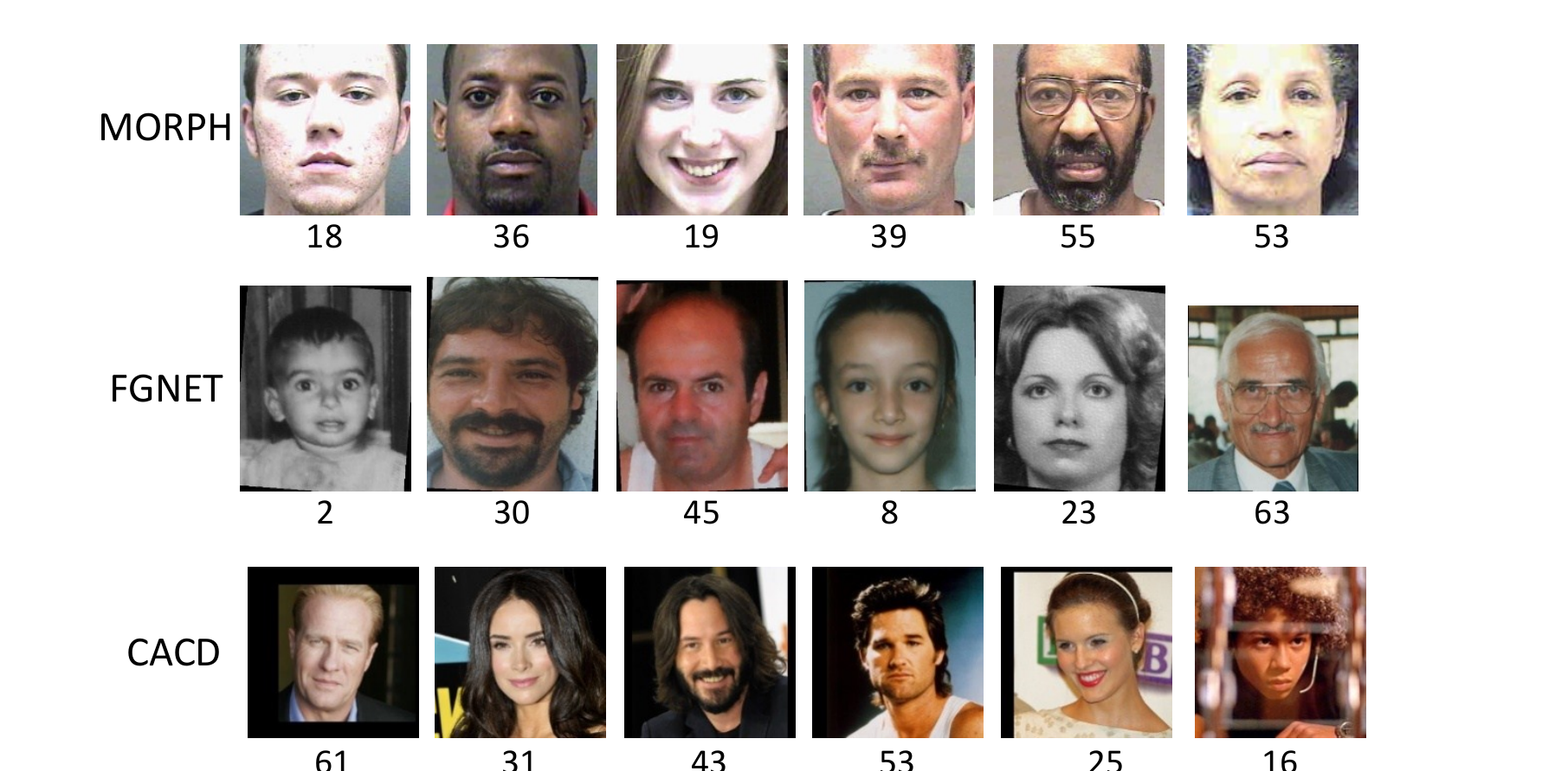}
\caption{Some examples of MORPH~\cite{Ref:MORPH06}, FG-NET~\cite{Ref:PanisLTC16} and CACD~\cite{Ref:ChenCH15}. The number below each image is the chronological age of each subject.}\label{fig:Example_Dataset}
\end{figure}

In this section we compare our DRFs with other state-of-the-art age estimation methods on three standard benchmarks: MORPH~\cite{Ref:MORPH06}, FG-NET~\cite{Ref:PanisLTC16} and the Cross-Age Celebrity Dataset (CACD)~\cite{Ref:ChenCH15}. Some examples of these three datasets are illustrated in Fig.~\ref{fig:Example_Dataset}.
%
%\begin{figure}[!t]
%\centering
%\includegraphics[width=1.0\linewidth]{databar.pdf}
%%\includegraphics[width=0.9\linewidth]{databar.pdf}
%\caption{Dataset statistics:
%(a) Age distribution of MORPH~\cite{Ref:MORPH06}, FG-NET~\cite{Ref:PanisLTC16} and CACD~\cite{Ref:ChenCH15};
%(b) Race and gender distribution of MORPH and CACD.}
%\label{fig:dset-stat}
%\end{figure}

\textbf{MORPH}
We first compare DRFs with other state-of-the-art age estimation methods on MORPH, which is the most popular dataset for age estimation.
MORPH contains more than 55,000 images from about 13,000 people of different races.
Each of the facial image is annotated with a chronological age.
The ethnicity of MORPH is very unbalanced, as more than 96\% of the facial
images are from African or European people.

Existing methods adopted different experimental settings on MORPH. The first setting (Setting I)
is introduced in~\cite{Ref:Chang11,Ref:Chen2013Cumulative,Ref:Guo08,Ref:Wang2015Deeply,Ref:Rothe2016Some,Ref:Rothe16,Ref:Agustsson17}, which selects 5,492 images of Caucasian Descent people
from the original MORPH dataset, to reduce the cross-ethnicity effects.
In Setting I, these 5,492 images are randomly partitioned into two subsets:
80\% of the images are selected for training and others for testing.
The random partition is repeated 5 times,
and the final performance is averaged over these 5 different partitions.
The second setting is used in~\cite{Ref:Geng2013Facial,Ref:Shen17,Ref:geng2016label,Ref:Gao2016Deep}, under which all of the images in MORPH are randomly split into training/testing ($80\%$/$20\%$) sets.
And also the random splitting is performed 5 times repeatedly. The final performance is obtained by averaging the performances of these 5 different splitting.
There are also several methods~\cite{Ref:GuoM11,Ref:guo2014framework,Ref:yi2014age} using the third setting (Setting III), which randomly selected a subset (about 21,000 images) from MORPH and restricted the ratio between Black and White and the one between Female and Male are 1:1 and 1:3, respectively.
For a fair comparison, we test the proposed DRFs on MORPH under all these three settings. The quantitative results of the three settings are summarized in Table.~\ref{table:morph-setting1}, Table.~\ref{table:morph-setting2} and Table.~\ref{table:morph-setting3}, respectively.
As can be seen from these tables, DRFs achieve the best performance on all of the settings, and outperform the current state-of-the-arts with a clear margin. There is only one method, dLDLF~\cite{Ref:Shen17}, which can achieve slightly worse result than DRFs (for setting II), as this method is also based on differentiable decision forests, but used for label distribution learning.

\begin{table}
\begin{center}
\begin{tabular}{l|c|c}
\hline
Method & MAE & CS  \\
\hline\hline
Human workers ~\cite{Ref:HanOLJ15} & 6.30 & 51.0 \%*\\
AGES ~\cite{Ref:geng2007automatic} & 8.83& 46.8 \%*\\
MTWGP ~\cite{Ref:zhang2010multi} & 6.28 & 52.1\%*\\
CA-SVR ~\cite{Ref:Chen2013Cumulative} & 5.88& 57.9\% \\
SVR ~\cite{Ref:Guo08} & 5.77 & 57.1\%\\
OHRank ~\cite{Ref:Chang11} & 6.07& 56.3\%\\
DLA ~\cite{Ref:Wang2015Deeply} & 4.77 & 63.4 \%*\\
Rank ~\cite{Ref:chang2010ranking} & 6.49& 49.1\%*\\
Rothe \etal~\cite{Ref:Rothe2016Some} & 3.45&  N/A\\
DEX ~\cite{Ref:Rothe16} & 3.25 &N/A\\
ARN ~\cite{Ref:Agustsson17} & 3.00&N/A\\
\hline
\textbf{DRFs(ours)} &\textbf{2.91} & \textbf{82.9\%}\\
\hline
\end{tabular}
\end{center}
\caption{Performance comparison on MORPH~\cite{Ref:MORPH06} (Setting I)(*: the value is read from the reported CS curve).}
\label{table:morph-setting1}
\end{table}

\begin{table}
\begin{center}
\begin{tabular}{l|c|c}
\hline
Method & MAE& CS  \\
\hline\hline
IIS-LDL ~\cite{Ref:Geng2013Facial} & 5.67&71.2\%*\\
CPNN ~\cite{Ref:Geng13} & 4.87& N/A\\
Huerta \etal~\cite{Ref:huerta2014facial} & 4.25 &71.2\%\\
BFGS-LDL ~\cite{Ref:geng2016label} & 3.94& N/A\\
OHRank ~\cite{Ref:Chang11} & 3.82&N/A\\
OR-SVM ~\cite{Ref:chang2010ranking} & 4.21&68.1\%*\\
CCA ~\cite{Ref:guo2013joint}  & 4.73&60.5\%*\\
LSVR ~\cite{Ref:GuoMFH09}  & 4.31&66.2\%*\\
OR-CNN ~\cite{Ref:Niu16} & 3.27& 73.0\%*\\
SMMR ~\cite{Ref:Huang17} & 3.24& N/A\\
Ranking-CNN ~\cite{Ref:ChenZDLR17} & 2.96 & 85.0\%*\\
DLDL ~\cite{Ref:Gao2016Deep} & 2.42& N/A\\
dLDLF ~\cite{Ref:Shen17} & 2.24& N/A\\
\hline
\textbf{DRFs(ours)} &\textbf{2.17}&\textbf{91.3\%}\\
\hline
\end{tabular}
\end{center}
\caption{Performance comparison on MORPH~\cite{Ref:MORPH06} (Setting II)(*: the value is read from the reported CS curve).}
\label{table:morph-setting2}
\end{table}

\begin{table}
\begin{center}
\begin{tabular}{l|l}
\hline
Method & MAE\\
\hline\hline
KPLS ~\cite{Ref:GuoM11} & 4.18 \\
Guo and Mu ~\cite{Ref:guo2014framework} & 3.92\\
CPLF~\cite{Ref:yi2014age} & 3.63 \\
\hline
\textbf{DRFs(ours)} &\textbf{2.98}\\
\hline
\end{tabular}
\end{center}
\caption{Performance comparison on MORPH~\cite{Ref:MORPH06} (Setting III).}
\label{table:morph-setting3}
\end{table}

%

%\begin{figure*}[!t]
%\centering
%\begin{tabular}{@{}ccc@{}}
%\includegraphics[width=0.5\linewidth]{ages-hist.pdf}
%\includegraphics[width=0.5\linewidth]{databar}
%\end{tabular}
%\end{figure*}

%\begin{figure}[!t]
%\centering
%\includegraphics[width=0.9\linewidth]{databar}
%\caption{Racial attribute of Morph and CACD datasets.}\label{fig:race-stat}
%\end{figure}

\textbf{FG-NET}
We then conduct experiments on FG-NET~\cite{Ref:PanisLTC16}, a dataset also widely used for age estimation. It contains 1002 facial images of 82 individuals,
in which most of them are white people. Each individual in FG-NET has more than 10 photos taken at different ages. The images in FG-NET have a large variation in lighting conditions,
poses and expressions.

Following the experimental setting used in~\cite{Ref:Yan2007Learning,Ref:Guo08,Ref:ChangCH11,Ref:Chen2013Cumulative,Ref:Rothe16}, we perform ``leave one out'' cross validation on this dataset, i.e., we leave images of one person for testing and take the remaining images for training.
The quantitative comparisons on FG-NET dataset are shown in Table.\ref{table:fg-net}.
As can be seen, DRFs achieve the state-of-the-art result with 3.85 MAE.
Note that, it is the only method that has a MAE below 4.0.
The age distribution of FG-NET is strongly biased,
moreover, the ``leave one out'' cross validation policy further aggravates
the bias between the training set and the testing set.
The ability of overcoming the bias between training and testing sets indicates that the proposed
DRFs can handle heterogeneous data well.

\begin{table}[htb]
\begin{center}
\begin{tabular}{l|c|c}
\hline
Method & MAE & CS \\
\hline\hline
Human workers ~\cite{Ref:HanOLJ15} & 4.70 & 69.5\%* \\
Rank ~\cite{Ref:chang2010ranking} & 5.79& 66.5\%*\\
DIF ~\cite{Ref:HanOLJ15}  & 4.80 &74.3\%*\\
AGES ~\cite{Ref:geng2007automatic} & 6.77& 64.1\%*\\
IIS-LDL ~\cite{Ref:Geng2013Facial} & 5.77&N/A\\
CPNN ~\cite{Ref:Geng13} & 4.76&N/A\\
MTWGP ~\cite{Ref:zhang2010multi}  & 4.83& 72.3\%*\\
CA-SVR ~\cite{Ref:Chen2013Cumulative} & 4.67& 74.5\%\\
LARR ~\cite{Ref:Guo08}  & 5.07 & 68.9\%*\\
OHRank ~\cite{Ref:Chang11} & 4.48& 74.4\%\\
DLA ~\cite{Ref:Wang2015Deeply}  & 4.26 &N/A\\
CAM~\cite{Ref:luu2011contourlet} & 4.12 & 73.5\%*\\
Rothe \etal~\cite{Ref:Rothe2016Some}  & 5.01& N/A\\
DEX ~\cite{Ref:Rothe16}  & 4.63&N/A\\
\hline
\textbf{DRFs (Ours)} &\textbf{3.85} & \textbf{80.6\%}\\
\hline
\end{tabular}
\end{center}
\caption{Performance comparison on FG-NET~\cite{Ref:PanisLTC16}(*: the value is read from the reported CS curve).}
\label{table:fg-net}
\end{table}

\textbf{CACD}
CACD~\cite{Ref:ChenCH15} is a large dataset which has around 160,000 facial images of 2,000 celebrities.
%
%200 of the celebrities are manually annotated, they are divided into two subset: the validation subset with 80 celebrities, and the testing subset with 120 celebrities.
%%
%The other 1200 celebrities compose the training set.
%
These celebrities are divided into three subsets: the training set which is composed of 1,800 celebrities, the testing set that has 120 celebrities and the validation set containing 80 celebrities.
Following~\cite{Ref:Rothe16}, we evaluate the performance of the models trained
on the training set and the validation set, respectively. The detailed comparisons are shown in Table~.\ref{table:cacd}.
The proposed DRFs model performs better than the competitor DEX~\cite{Ref:Rothe16}, no matter which set they are trained on. It's worth noting that, the improvement of DRFs to DEX is much more significant when they are trained on the validation set than the training set. This result can be explained in this way: As we described earlier, the heterogeneous data is the main challenge in training age estimation models. This challenge can be alleviated by enlarging the scale of training data. Therefore, DEX and our DRFs achieve comparable results when they are trained on the training set. But when they are trained on the validation set, which is much smaller than the training set, DRFs outperform DEX significantly, because we directly address the inhomogeneity challenge. Therefore, DRFs are capable of handling heterogeneous data even learned from a small set.

\begin{table}
\begin{center}
\begin{tabular}{l|c|c}
\hline
Trained on & Dex~\cite{Ref:Rothe16} & \textbf{DRFs (Ours)} \\
\hline\hline
CACD (train)  & 4.785 & \textbf{4.637}\\
CACD (val) &  6.521& \textbf{5.768}\\
\hline
\end{tabular}
\end{center}
\caption{Performance comparison on CACD (measured by MAE)~\cite{Ref:ChenCH15}.}
\label{table:cacd}
\end{table}
\subsection{Discussion}
\subsubsection{Visualization of Learned Leaf Nodes}
\begin{figure}[!t]
\centering
\includegraphics[trim=0cm 6cm 0cm 5cm, clip=true, width=1.0\linewidth]{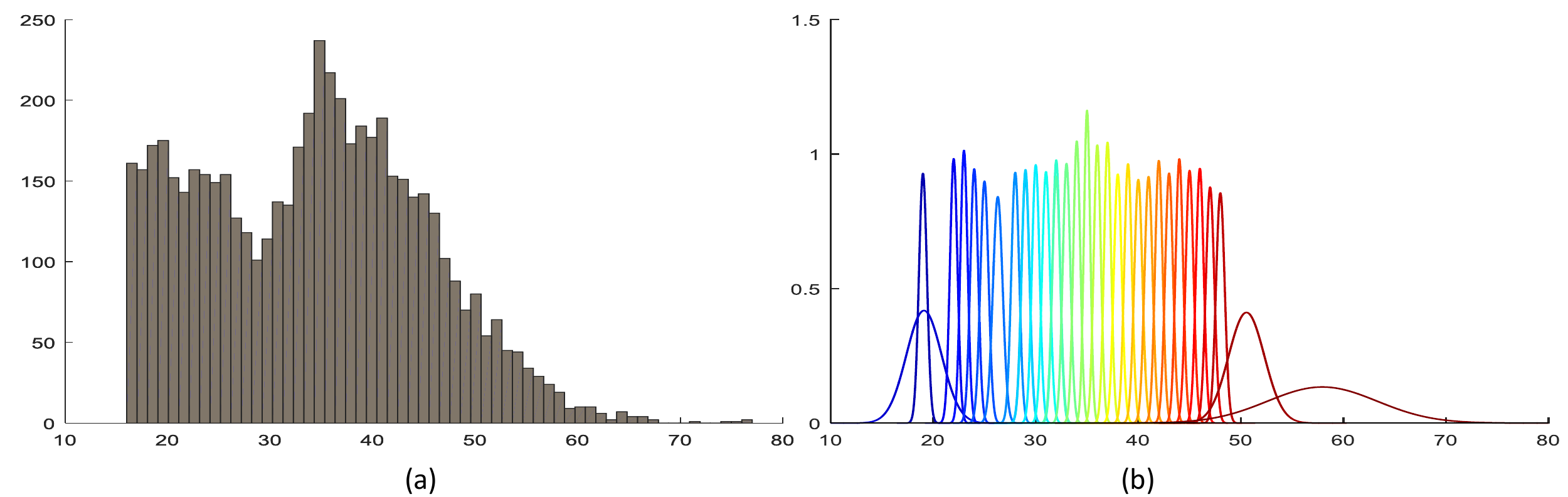}
\caption{(a) Histogram of data samples with respect to age on MORPH~\cite{Ref:MORPH06} (Setting I). (b) Visualization of the learned leaf node distributions in our DRFs (best viewed in color).}\label{fig:LeafNodeDistr}
\end{figure}

To better understand DRFs, we visualize the distributions at leaf nodes learned on MORPH~\cite{Ref:MORPH06} (Setting I) in Fig.~\ref{fig:LeafNodeDistr}(b). Each leaf node contains a Gaussian distribution (the vertical and horizontal axes represent probability density and age, respectively). For reference, we also display the histogram of data samples (the vertical axis) with respect to age (the horizontal axis). Observed that, the mixture of these Gaussian distributions learned at leaf nodes is very similar to the histogram of data samples, which indicates our DRFs fit the age data well. The age data in MORPH was sampled mostly below age 60, and densely concentrated around 20's and 40's. So the Gaussian distribution centered around 60 has much larger variance than those centered in the interval between 20 and 50, but has smaller probability density. This is because although these learned Gaussian distributions represent homogeneous local partitions, the number of samples is not necessarily uniformly distributed among partitions. Another phenomenon is these Gaussian distributions are heavily overlapped, which accords with the fact that different people with the same age but have quite different facial appearances.
\subsubsection{Parameter Discussion}
\begin{figure}[!t]
\centering
\includegraphics[trim=0cm 6cm 0cm 4cm, clip=true, width=1.0\linewidth]{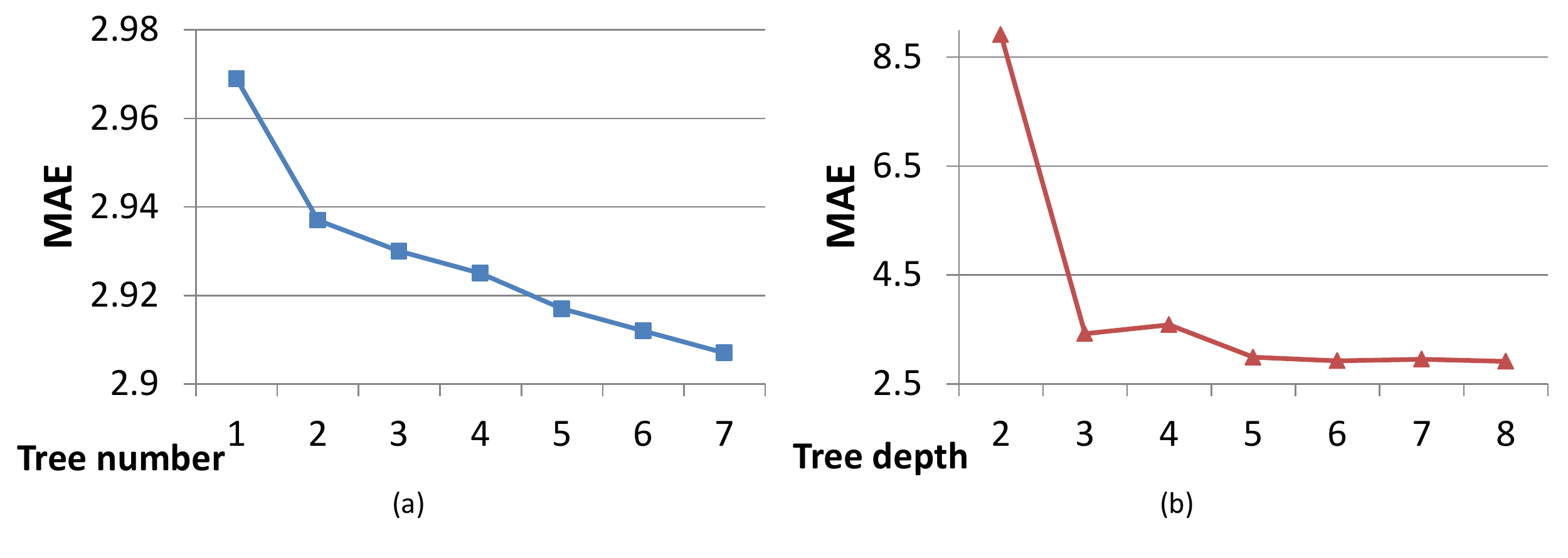}
\caption{Performance changes by varying (a) tree number and (b) tree depth on MORPH~\cite{Ref:MORPH06} (Setting I).}\label{fig:ParameterDiscussion}
\end{figure}

The tree number and tree depth are two important hyper-parameters for our DRFs. Now we vary each of them and fix the other one to the default value to see how the performance changes on MORPH (Setting I). As shown in Fig.~\ref{fig:ParameterDiscussion}, using more trees leads to a better performance as we expected, and with the tree depth increase, the MAE first becomes lower and then stable.

\section{Conclusion}
We proposed Deep Regression Forests (DRFs) for age estimation, which learn nonlinear regression between heterogeneous facial feature space and ages. In DRFs, by performing soft data partition at split nodes, the forests can be connected to a deep network and learned in an end-to-end manner, where data partition at split nodes is learned by Back-propagation and data abstraction at leaf nodes is optimized by iterating a step-size free and fast-converged update rule derived from Variational Bounding. The end-to-end learning of split and leaf nodes ensures that partition function at each split node is input-dependent and the local input-output correlation at each leaf node is homogeneous. Experimental results showed that DRFs achieved state-of-the-art results on three age estimation benchmarks.
{\small
\bibliographystyle{ieee}
\bibliography{mybib}
}

\end{document}